\documentclass{article}
\usepackage{spconf,amsmath,graphicx}
 
\usepackage[]{graphicx}
\graphicspath{{Figures/}}
\usepackage{caption}
\usepackage{subfigure}
\usepackage{amsmath}
\usepackage{amssymb}
\usepackage{epsfig}
\usepackage{cite}
\usepackage{color}
\usepackage{balance}
\usepackage{pgfplots}

\usepackage{amsmath}
\usepackage{amsfonts} 
\usepackage{graphicx}
\usepackage{pdfpages}
\usepackage[T1]{fontenc} 
\usepackage{array}
\usepackage{url}

\usepackage{color}
\usepackage{wrapfig}
\usepackage{lipsum}
\usepackage[hidelinks]{hyperref}
\usepackage{multirow}
\usepackage{tikz}

\usepackage{tabularx}

\title{Nighttime sky/cloud image segmentation}

\name{Soumyabrata Dev,$^{1}$ Florian M. Savoy,$^{2}$ Yee Hui Lee,$^{1}$ Stefan Winkler$^{2}$~\thanks{This research is funded by the Defence Science and Technology Agency (DSTA), Singapore.}~\thanks{Send correspondence to \url{stefan.winkler@adsc.com.sg}.}}
\address{$^{1}$~School of Electrical and Electronic Engineering, Nanyang Technological University (NTU), Singapore \\ 
$^{2}$~Advanced Digital Sciences Center (ADSC), University of Illinois at Urbana-Champaign, Singapore 
}

\begin{document}

\maketitle

\begin{abstract}
Imaging the atmosphere using ground-based sky cameras is a popular approach to study various atmospheric phenomena. However, it usually focuses on the daytime. Nighttime sky/cloud images are darker and noisier, and thus harder to analyze. An accurate segmentation of sky/cloud images is already challenging because of the clouds' non-rigid structure and size, and the lower and less stable illumination of the night sky increases the difficulty. Nonetheless, nighttime cloud imaging is essential in certain applications, such as continuous weather analysis and satellite communication. 

In this paper, we propose a superpixel-based method to segment nighttime sky/cloud images.  We also release the first nighttime sky/cloud image segmentation database to the research community. The experimental results show the efficacy of our proposed algorithm for nighttime images.

\end{abstract}


\setlength{\fboxsep}{0pt}
\setlength{\fboxrule}{0.2pt}

\section{Introduction}
\label{sec:intro}

Over the last few decades, there has been growing interest in geoscience to study clouds and analyze their various features. 
Satellite images are the conventional source for analyzing the formation of clouds, their movement, and other atmospheric properties. However, these images suffer from poor spatial and/or temporal resolution. Therefore, remote sensing analysts increasingly turn to ground-based sky imagers~\cite{GRSM2016} that are able to produce images with significantly higher spatial resolution at much more frequent intervals. An accurate segmentation of the cloud mass from these images is of prime importance for any kind of cloud analysis. 

Almost all existing works in sky/cloud image segmentation use color as the discriminatory feature to detect clouds in ground-based sky camera images. This is because the sky is predominantly blue during daytime owing to Rayleigh scattering of light. However, this assumption no longer holds during nighttime.  

In the literature, most existing works consider only daytime images. Several color models are used in cloud segmentation~\cite{GRSL2017,Li2011,ICIP1_2014,ICIP2015a}. Yang et al.\ (2009) used the Otsu threshold on red-blue difference images~\cite{Yang2009cloud}. Yang et al.\ (2010) employed a local method~\cite{Yang2010} for cloud segmentation, which uses an adaptive threshold on sub-images of the sky. Recently, Liu et al.\ \cite{LiuSP2015} used a superpixel approach and generated corresponding thresholds for each superpixel to compute the final binary image. The local thresholds of these superpixels are empirically evaluated based on a set of case-based decisions, thereby generating a matrix of thresholds.  

Cloud segmentation in the nighttime sky is often neglected. Yet, analyzing clouds at nighttime helps in several areas viz.\ weather reporting and prediction, aviation, and satellite communication. Only very recently, Gacal et al.\ in ~\cite{Gacal2016} explored the nighttime cloud detection using ground-based sky cameras. Their method uses a constant fixed threshold in the gray-scale channel of the cloud images, captured over Manila region in the Philippines. 

Addressing this gap in the literature, we propose an efficient nighttime cloud segmentation approach. In this paper, we perform a careful selection of color spaces and components, and thereby propose a superpixel-based approach to efficiently segment nighttime sky/cloud image. We also present a new nighttime sky/cloud image segmentation database, which we use for benchmarking our approach.

\begin{figure*}[htb]
\centering
\fbox{\includegraphics[width=0.19\textwidth]{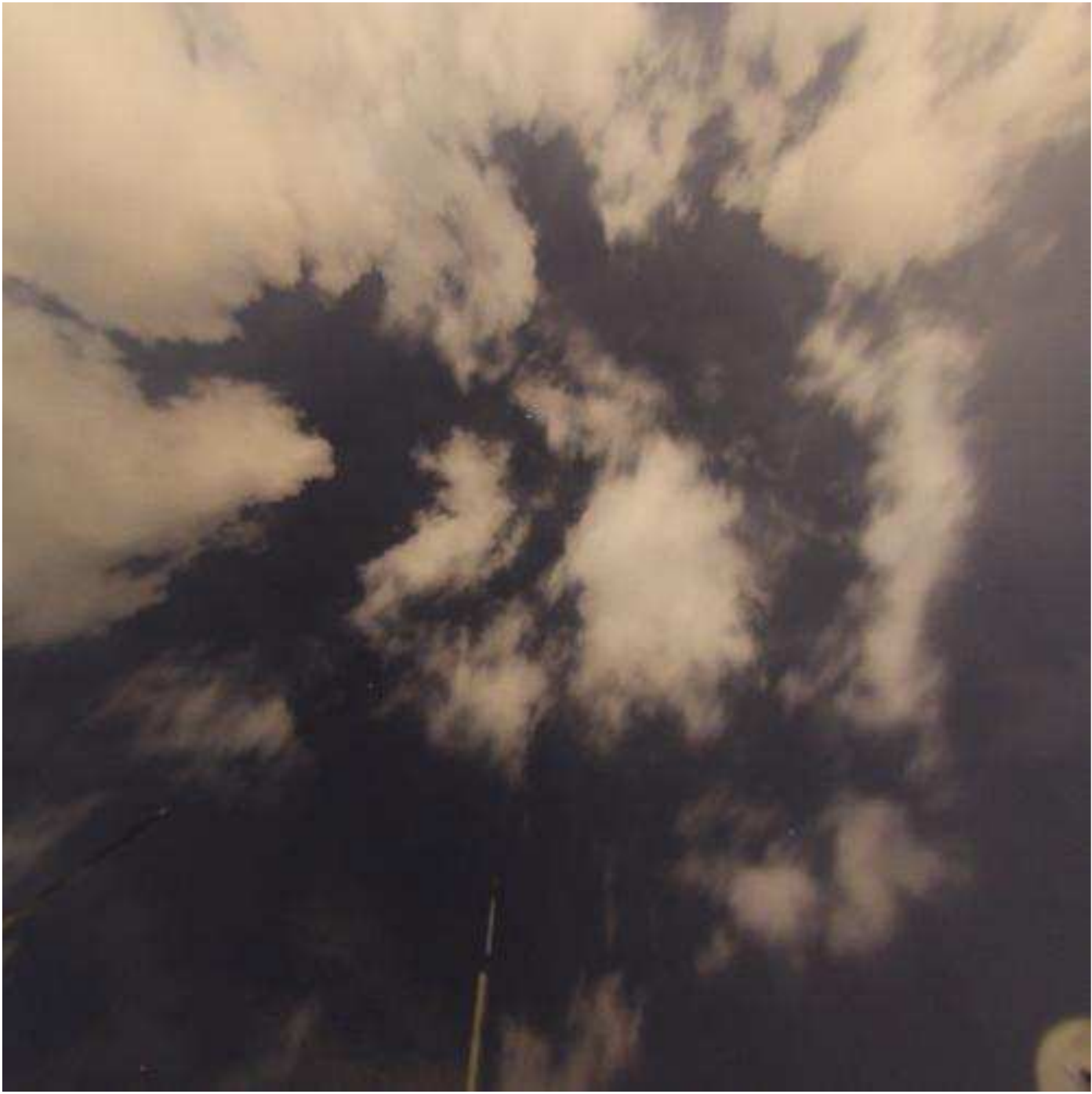}}\hspace{0.5mm}    
\fbox{\includegraphics[width=0.19\textwidth]{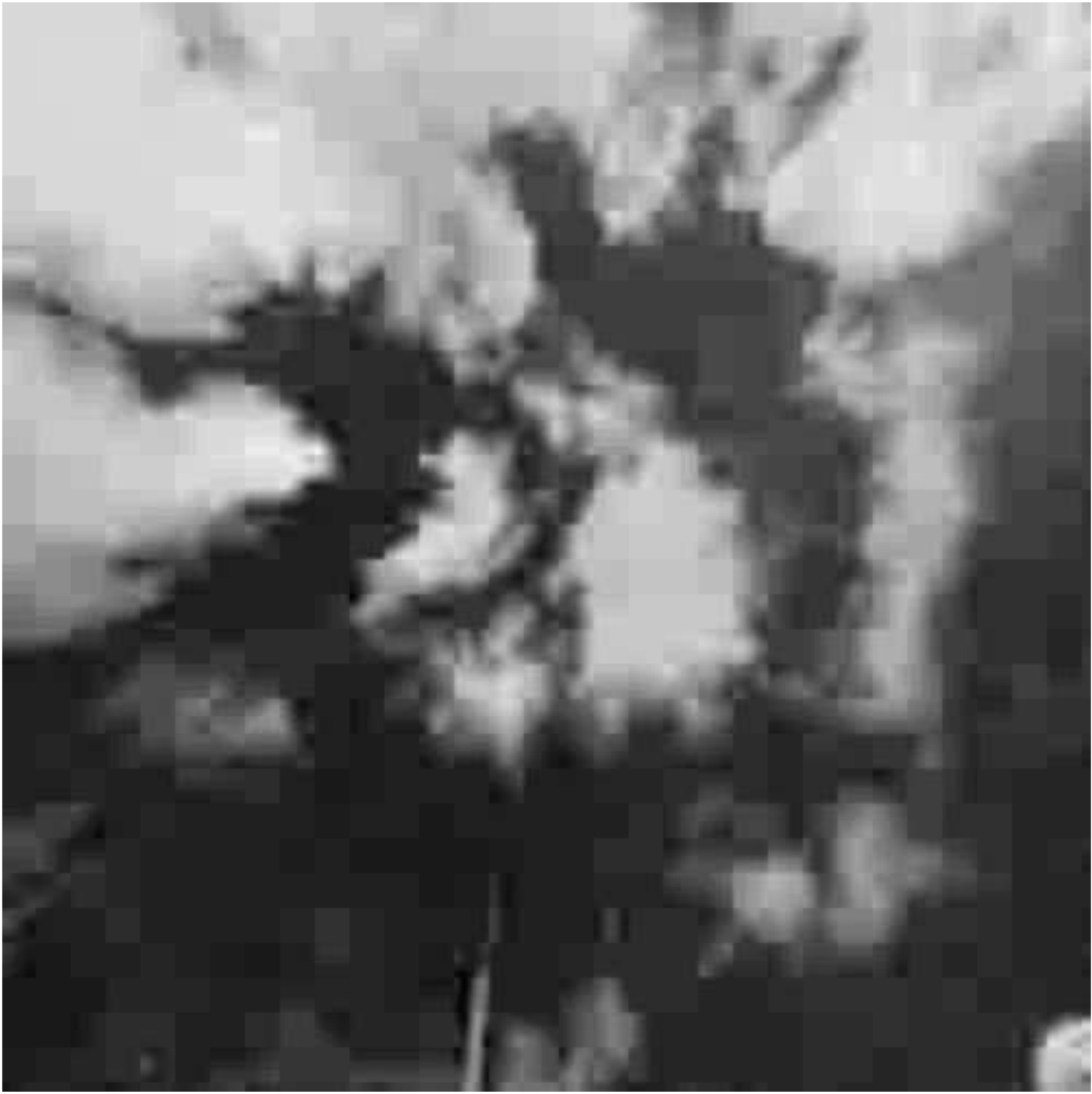}}\hspace{0.5mm} 
\fbox{\includegraphics[width=0.19\textwidth]{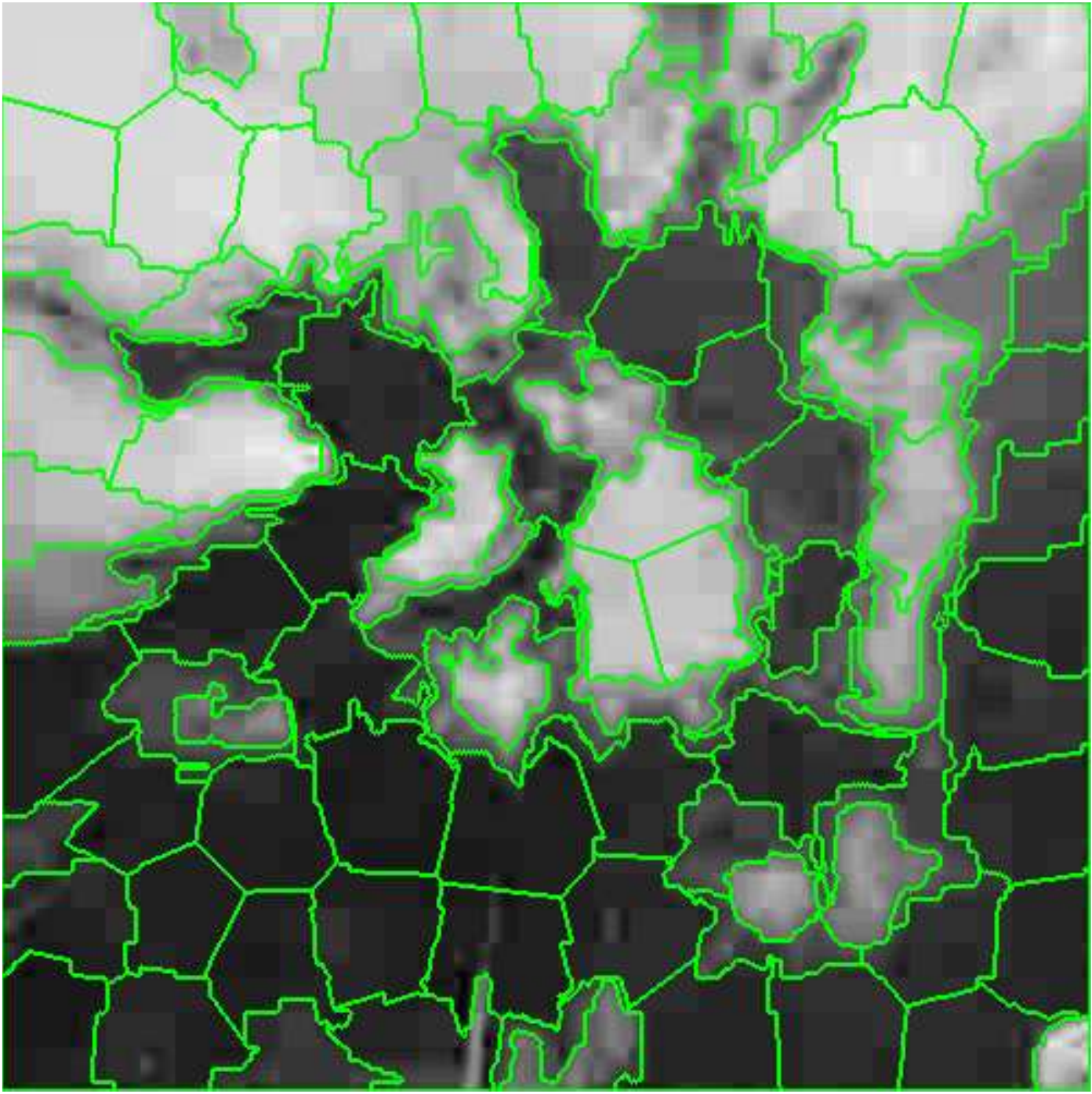}}\hspace{0.5mm}
\fbox{\includegraphics[width=0.19\textwidth]{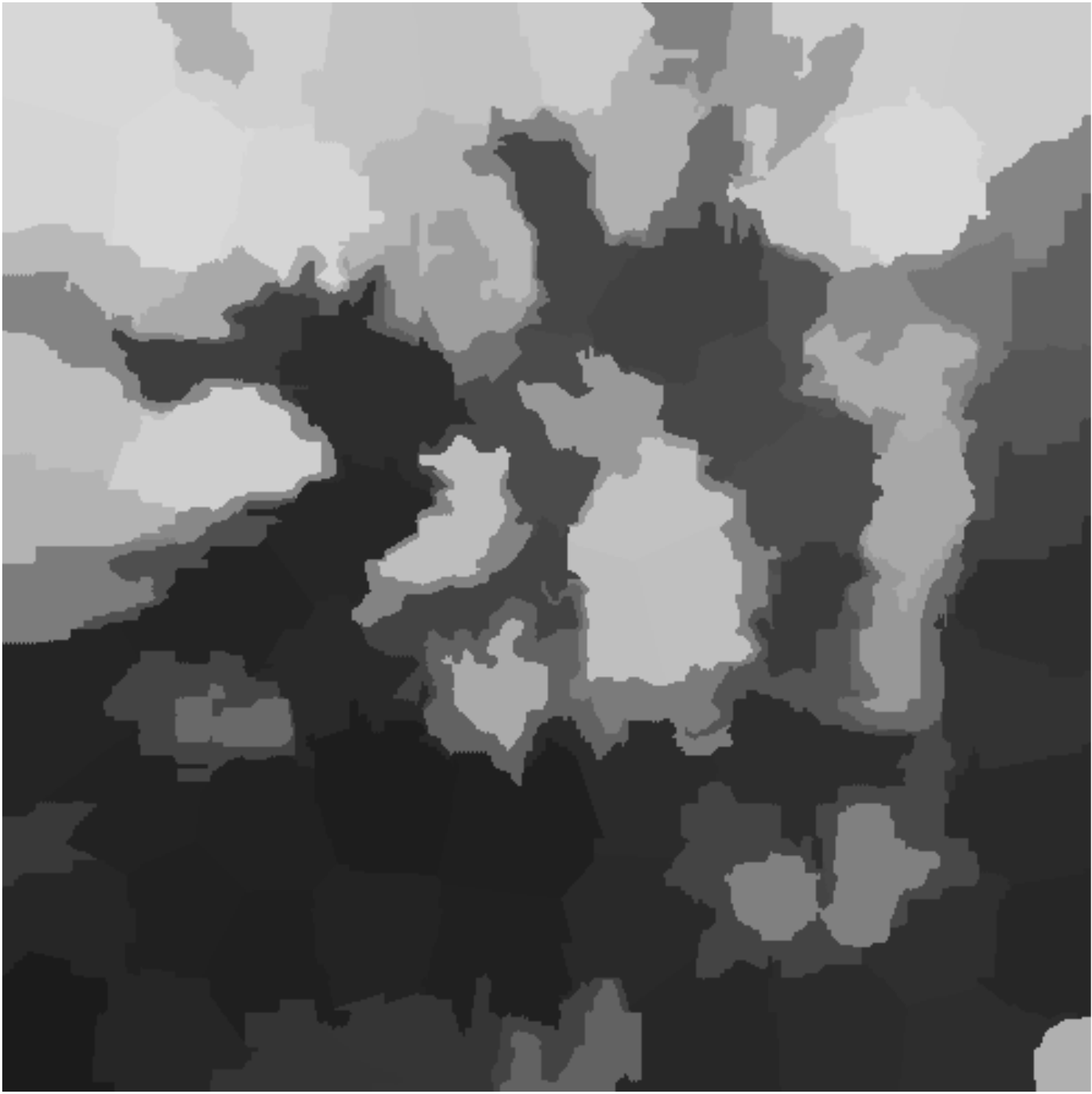}}\hspace{0.5mm}
\fbox{\includegraphics[width=0.19\textwidth]{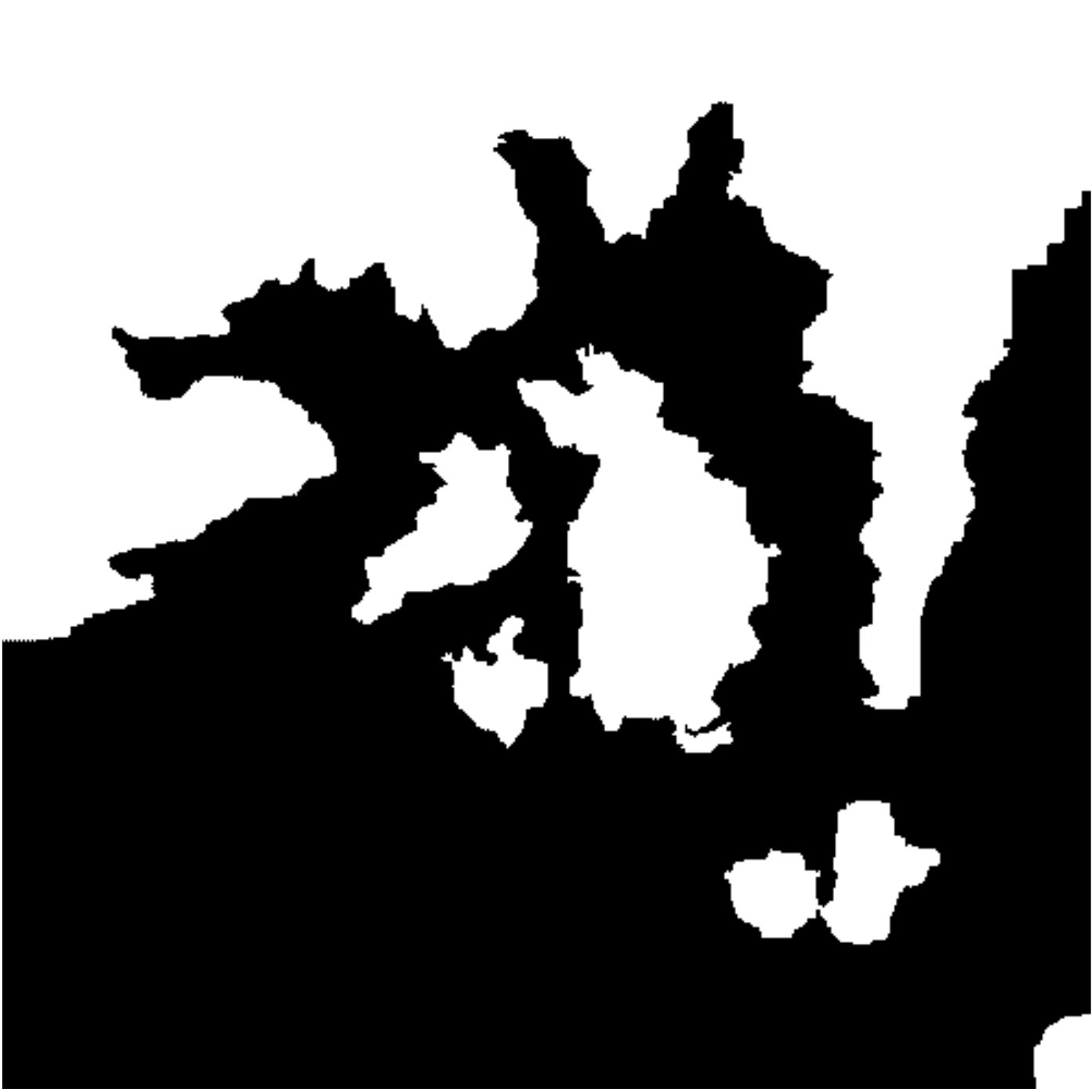}}\hspace{0.5mm}\\
\makebox[0.19\textwidth][c]{(a)}
\makebox[0.19\textwidth][c]{(b)}
\makebox[0.19\textwidth][c]{(c)}
\makebox[0.19\textwidth][c]{(d)}
\makebox[0.19\textwidth][c]{(e)}
\caption{Proposed superpixel-based nighttime sky/cloud image segmentation approach. (a) Input image; (b) Red-blue difference image; (c) Over-segmentation using superpixels; (d) Indexed image using average red-blue difference of the superpixels; (e) Final sky/cloud segmentation result.}
\label{fig:night-story}
\end{figure*}

\section{Nighttime Cloud Segmentation}
\label{sec:main}

In this section, we present our detailed segmentation framework for nighttime cloud segmentation.\footnote{~The source code of all simulations in this paper is available online at \url{https://github.com/Soumyabrata/nighttime-imaging}.} Our approach is illustrated in Fig.~\ref{fig:night-story}. Using an input image as shown in Fig.~\ref{fig:night-story}(a), we obtain a suitable color channel (in this example the red-blue difference), shown in Fig.~\ref{fig:night-story}(b). We then perform an over-segmentation of this image using a superpixel technique; the over-segmented image is shown in Fig.~\ref{fig:night-story}(c). Thereafter, we convert it into an indexed image as shown in Fig.~\ref{fig:night-story}(d) using a quantization procedure. Subsequently, we cluster these superpixels into two clusters -- \emph{cloud} and \emph{sky}. The resulting binary segmented image is shown in Fig.~\ref{fig:night-story}(e).
 
\begin{figure}[htb]
\centering
\includegraphics[height=0.21\textwidth]{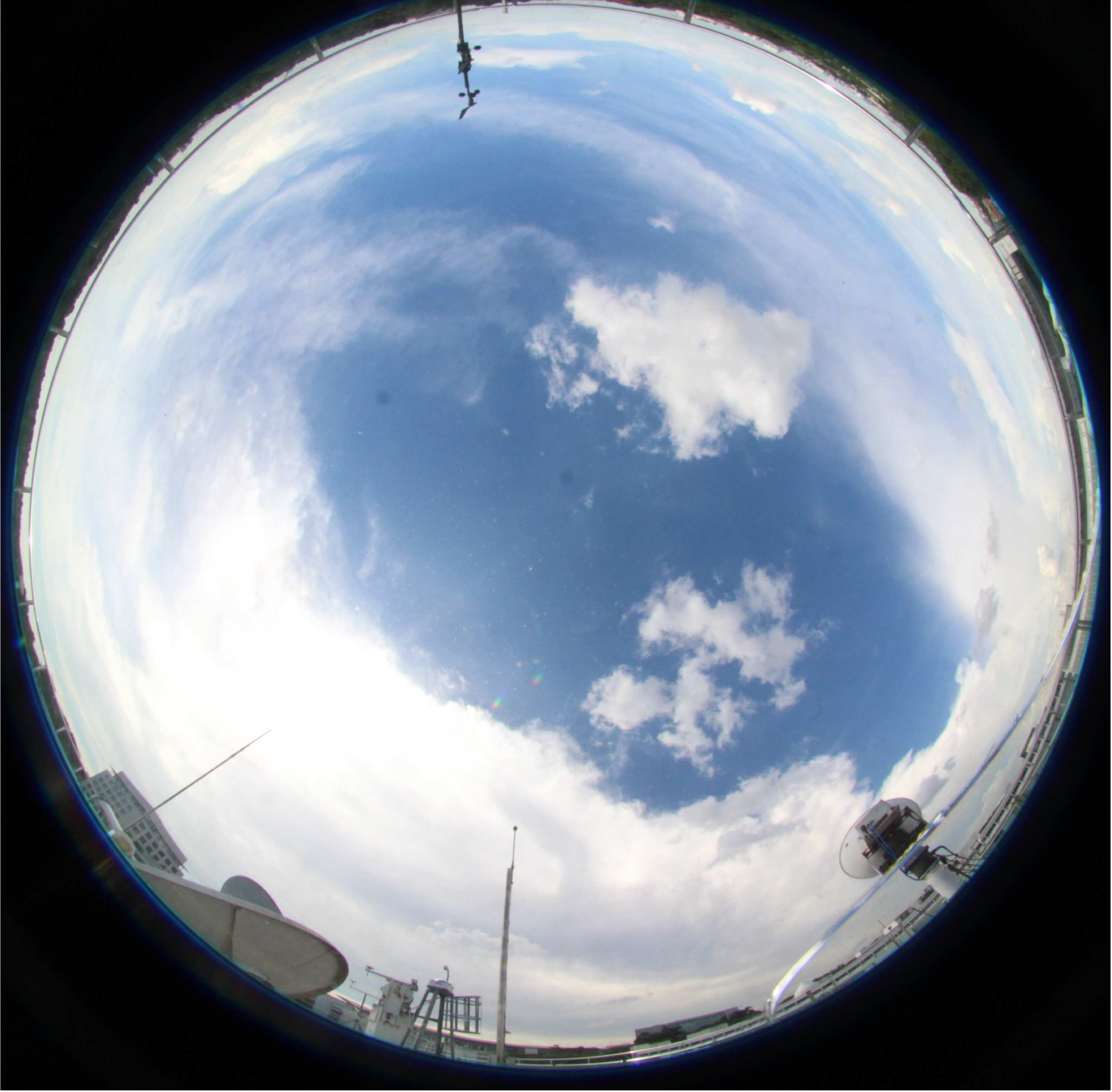}\hspace{0.5mm}    
\includegraphics[height=0.21\textwidth]{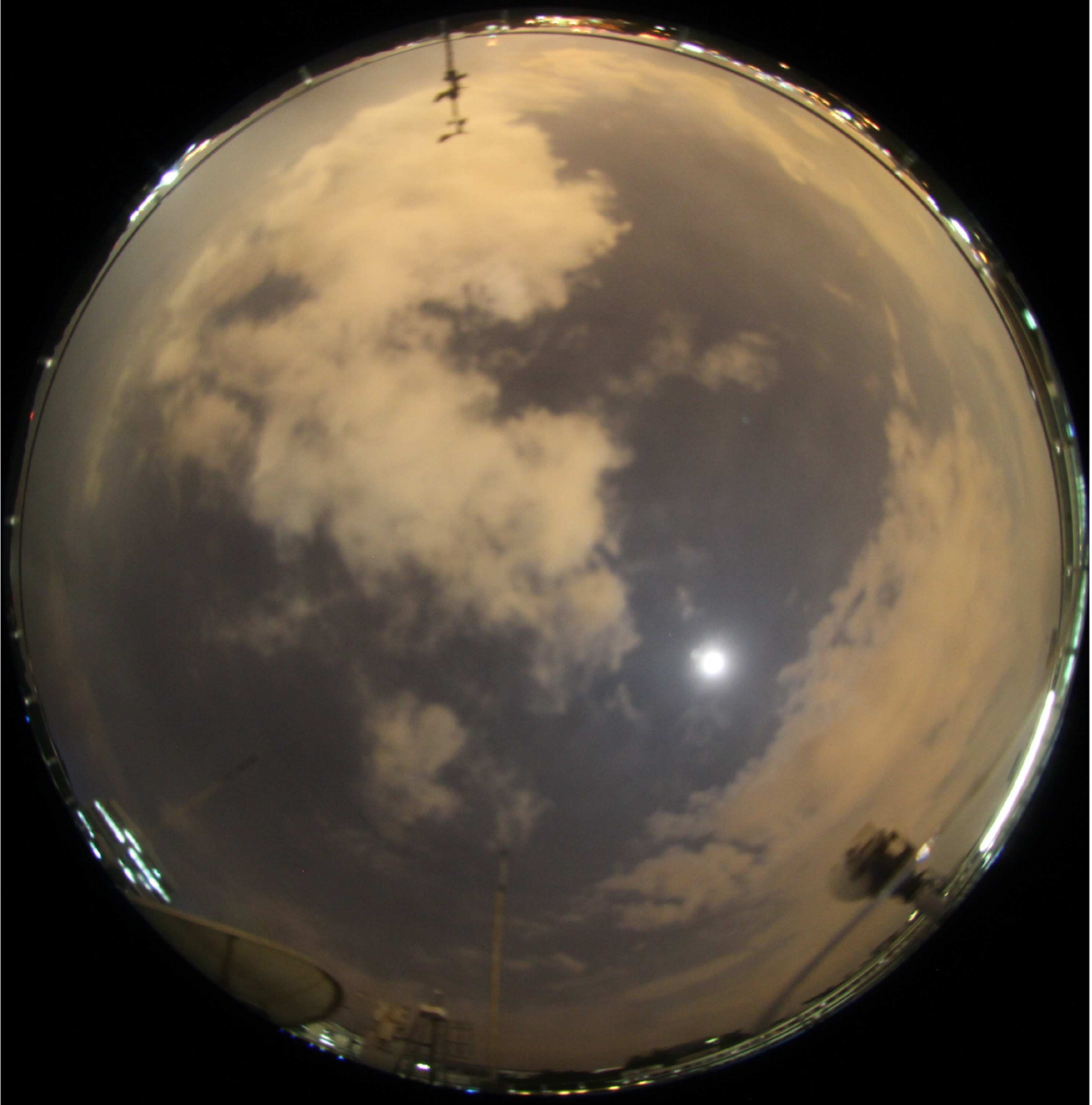}\\
\makebox[0.23\textwidth][c]{\small{15:34 -- 1/125s, ISO 400, f/22.6}}
\hspace{0.5mm} 
\makebox[0.23\textwidth][c]{\small{03:30 -- 2s, ISO 800, f/2.8}}
\caption{Sample day- and night-time image captured by our sky camera. Note the different camera settings (exposure time, ISO, f-number) in the two images.}
\label{fig:dnImages}
\end{figure}

Unlike the conventional daytime sky/cloud image, nighttime images include several additional imaging challenges. 
Figure~\ref{fig:dnImages} shows a sample day- and night-time image captured by our high resolution sky camera. Unlike the clear blue sky and white clouds seen during daytime, the night sky and clouds appear more reddish. This is mainly because of increasing light pollution in urban areas like Singapore. In a cloudy night-sky, the longer wavelengths of the artificial light are scattered back to the earth's surface, providing a reddish tinge to the night sky. More details on this phenomenon can be found in a recent study \cite{Kyba2012}. Moreover, nighttime images are more noisy and blurry in nature, with less distinctive features, owing to the higher exposure time and lower f-numbers of the camera lens.

\subsection{Notation}
Let us suppose that an image is represented by $\mathbf{X} \in {\rm I\!R}^{m \times n \times 3}$. We perform an analysis of different color spaces and components that are frequently used in cloud segmentation. Table~\ref{tab:color-analysis} shows the $16$ color channels, represented by $c_{1-16}$, that are used in our analysis. 

\begin{table}[htb]
\small
\centering
\setlength{\tabcolsep}{4pt} 
\begin{tabular}{c|c||c|c||c|c||c|c||c|c||c|c}
  \hline
  $c_{1}$ & R & $c_{4}$ & H & $c_{7}$ & Y & $c_{10}$ & $L^{*}$ & $c_{13}$ & $R/B$ & $c_{16}$ & $C$\\
  $c_{2}$ & G & $c_{5}$ & S & $c_{8}$ & I & $c_{11}$ & $a^{*}$ & $c_{14}$ & $R-B$& $ $ & $ $\\
  $c_{3}$ & B & $c_{6}$ & V & $c_{9}$ & Q & $c_{12}$ & $b^{*}$ & $c_{15}$ & $\frac{B-R}{B+R}$ & $ $ & $ $\\
  \hline
\end{tabular}
\caption{Color spaces and components used in our analysis of nighttime images.}
\label{tab:color-analysis}
\end{table}

It consists of channels from color models such as $RGB$, $HSV$, $YIQ$, $L^{*}a^{*}b^{*}$, various combinations of red and blue color channels, and chroma ($C$). We choose the most discriminatory color channel amongst these channels for our segmentation approach (see Section~\ref{sec:which-channel}). Let us denote this discriminatory channel by $\mathbf{c_{*}} \in {\rm I\!R}^{m \times n}$. We represent the feature vector used during superpixel clustering as $\mathbf{X}_f$, and the value of the channel $\mathbf{c_{*}}$ at the image co-ordinates $(x,y)$ is $r$. Finally, the number of superpixels is $P$. 

\subsection{Proposed Algorithm}
As described in Fig.~\ref{fig:dnImages}, the nighttime images are noisy in nature, and pixel-wise classification into day- and night- pixel is not conducive. Therefore, we employ a superpixel-based method that groups \emph{similar} pixels together in a single superpixel, while respecting cloud boundaries. We use the superpixel generation algorithm, Simple Linear Iterative Clustering (SLIC)~\cite{SLIC} in our proposed nighttime image segmentation approach. Instead of the default CIE LAB color space used by SLIC, we use the discriminatory channel $\mathbf{c_{*}}$ for the superpixel clustering. The clustering is performed on the $3D$ feature space $\mathbf{X}_f$. We calculate this feature as:
\begin{equation}
\label{eq:feature}
\begin{aligned}
\mathbf{X}_f=[\mathbf{r}  \, \mathbf{x} \, \mathbf{y}],
\end{aligned}
\end{equation}
where the vectors $\mathbf{r}$, $\mathbf{x}$, $\mathbf{y}$ represent the value, x- and y-coordinate for the vectorized form of image $\mathbf{c_{*}}$. This $3D$ feature is used for superpixel clustering, similar to the $5D$ LAB clustering technique used in SLIC approach~\cite{SLIC}.  We thereby obtain an over-segmented image on the channel $\mathbf{c_{*}}$. An example is illustrated in Fig.~\ref{fig:night-story}(c).

The generated superpixels in Fig.~\ref{fig:night-story}(c) respect the sky/cloud image boundaries very well, under the assumption that each superpixel contains only \emph{sky} or only \emph{cloud} pixels. Therefore, we compute the indexed image by mapping each pixels inside a superpixel to a common value $\bar{r}$. This common quantity $\bar{r}$ for a superpixel is estimated by averaging all the $r$ values in that corresponding superpixel. The indexed image is shown in Fig.~\ref{fig:night-story}(d). This vector quantization in the color channel $\mathbf{c_{*}}$ is performed to reduce the color depth of the discriminatory channel, and is inspired by Puzicha et al.\ \cite{Puzicha2000}. 

Finally, we perform an unsupervised k-means clustering on this quantized channel, whose feature vector comprises the average value $\bar{r}$ for the individual superpixels. We perform this clustering into \emph{sky} and \emph{cloud} clusters. The cluster center with lower value is assigned \emph{sky} label, and the one with higher value is assigned \emph{cloud} label. The final binary mask obtained via k-means clustering is illustrated in Fig.~\ref{fig:night-story}(e).

In our proposed approach, the only input parameter is the number of superpixels $P$ used in the channel $\mathbf{c_{*}}$, where $P \in \mathbb{N}, 1 \leq T \leq mn$. Therefore, the approximate size of a superpixel is $(mn/P)$ pixels. The constant $P$ is an indication of the degree of over-segmentation of the image. A large $P$ facilitates capturing fine features of the input image. We should set $P$ to a large enough value, such that each superpixel contain \emph{only} sky or cloud pixels. More discussion on the value of $P$ can be found in Section~\ref{sec:illust}.

\section{Database}
\label{sec:print}
While there has been extensive work on daytime sky/cloud image segmentation, nighttime sky imaging is rare. Therefore, there are no publicly available nighttime sky/cloud image segmentation databases. We therefore created a new database specifically for this purpose. In analogy to our popular SWIMSEG database with daytime sky images published earlier \cite{JSTARS2016}, we refer to this database as Singapore Whole sky Nighttime Imaging SEGmentation Database (SWINSEG).\footnote{~The SWINSEG dataset is available for download at \url{http://vintage.winklerbros.net/swinseg.html}.}

\begin{figure}[htb]
\centering
\includegraphics[width=0.23\columnwidth]{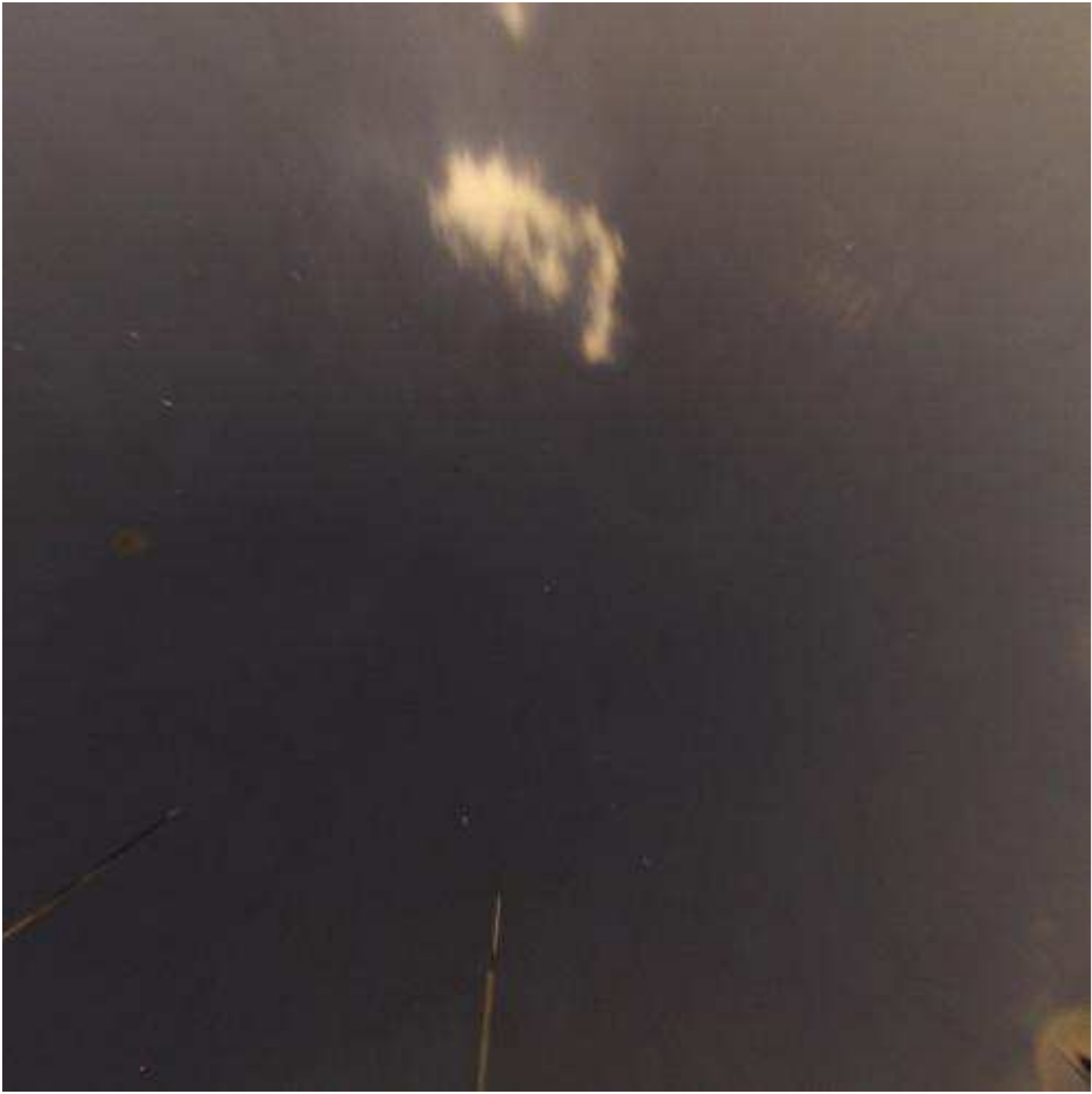} 
\includegraphics[width=0.23\columnwidth]{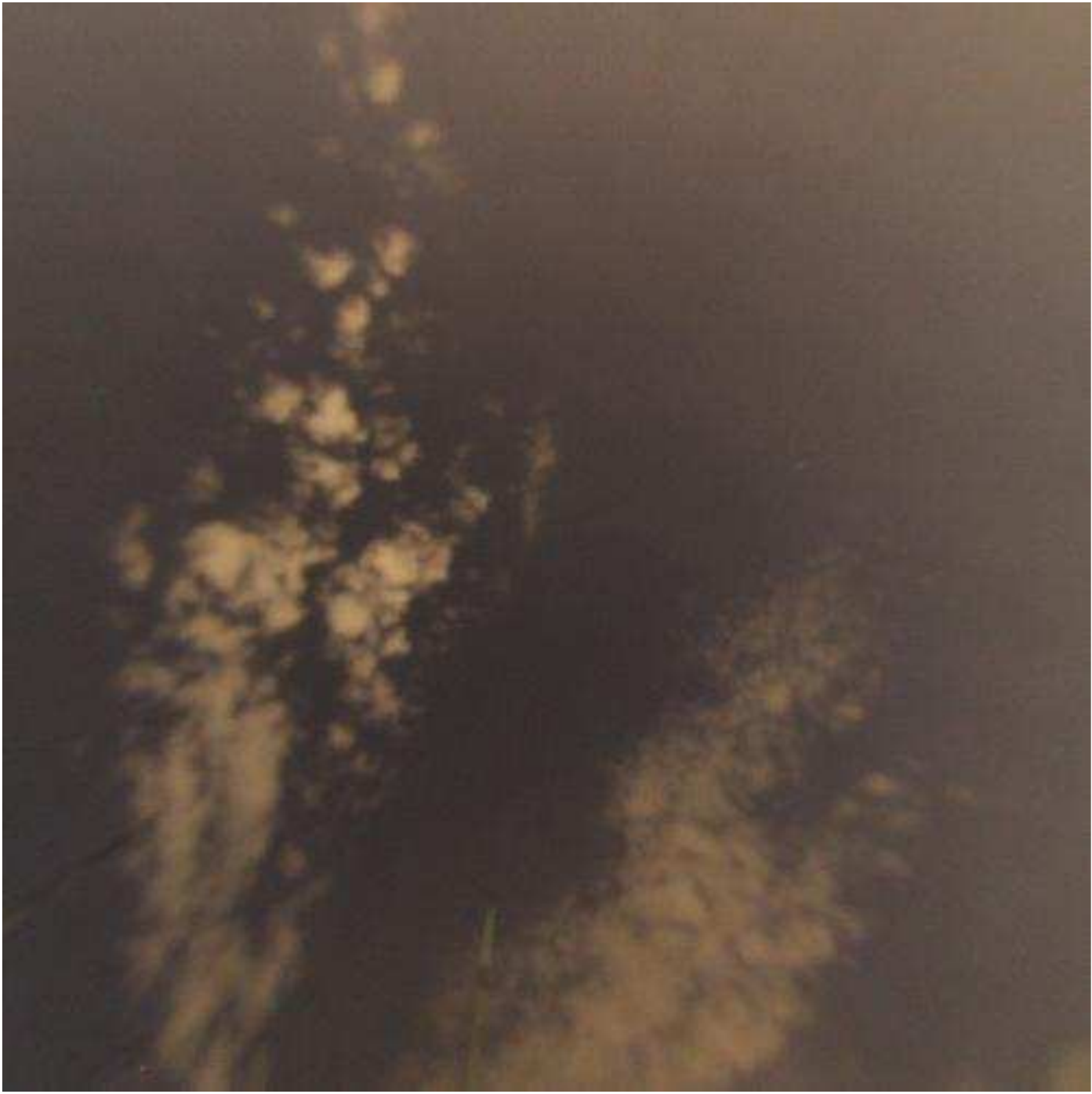}  
\includegraphics[width=0.23\columnwidth]{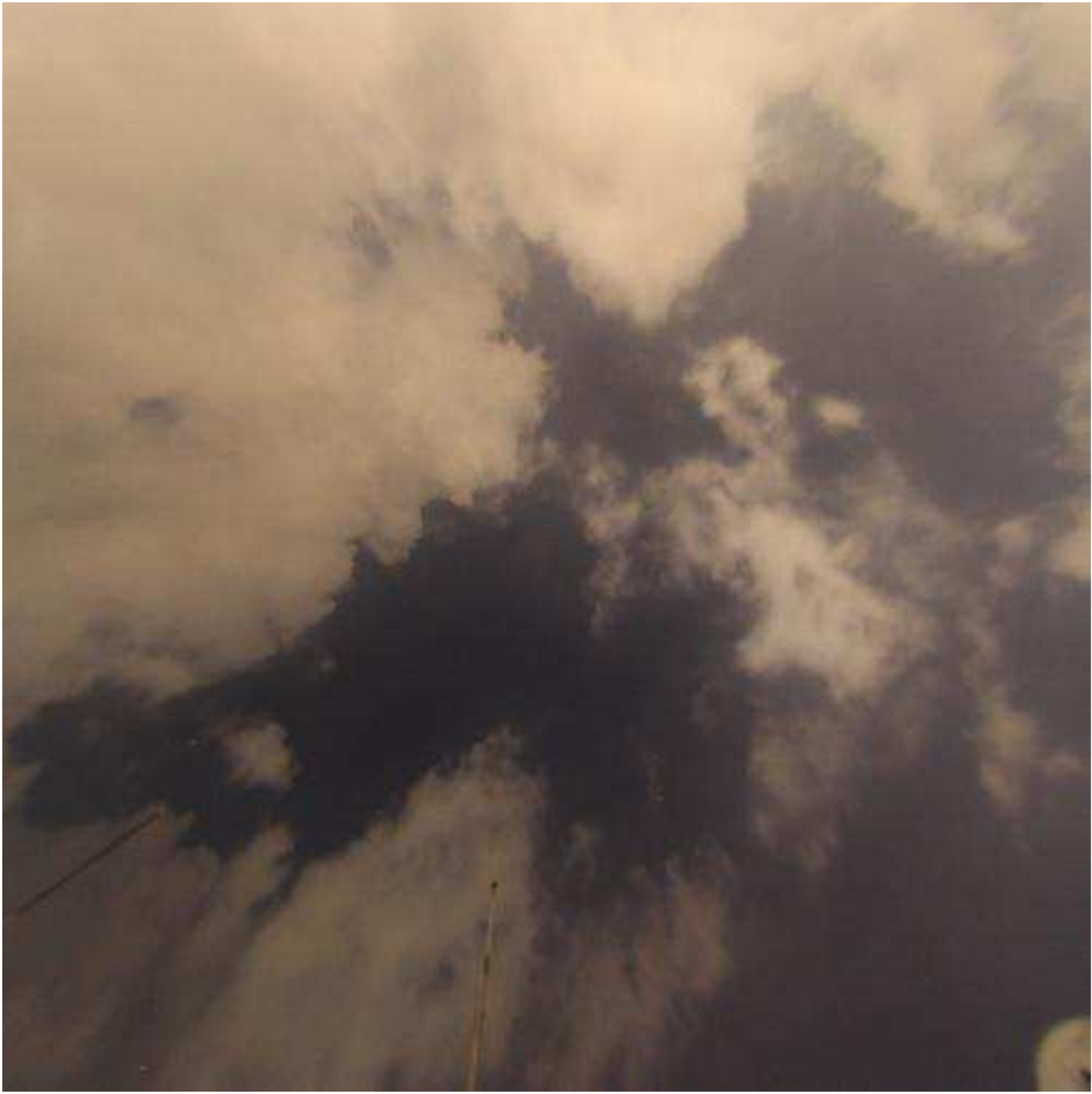}
\includegraphics[width=0.23\columnwidth]{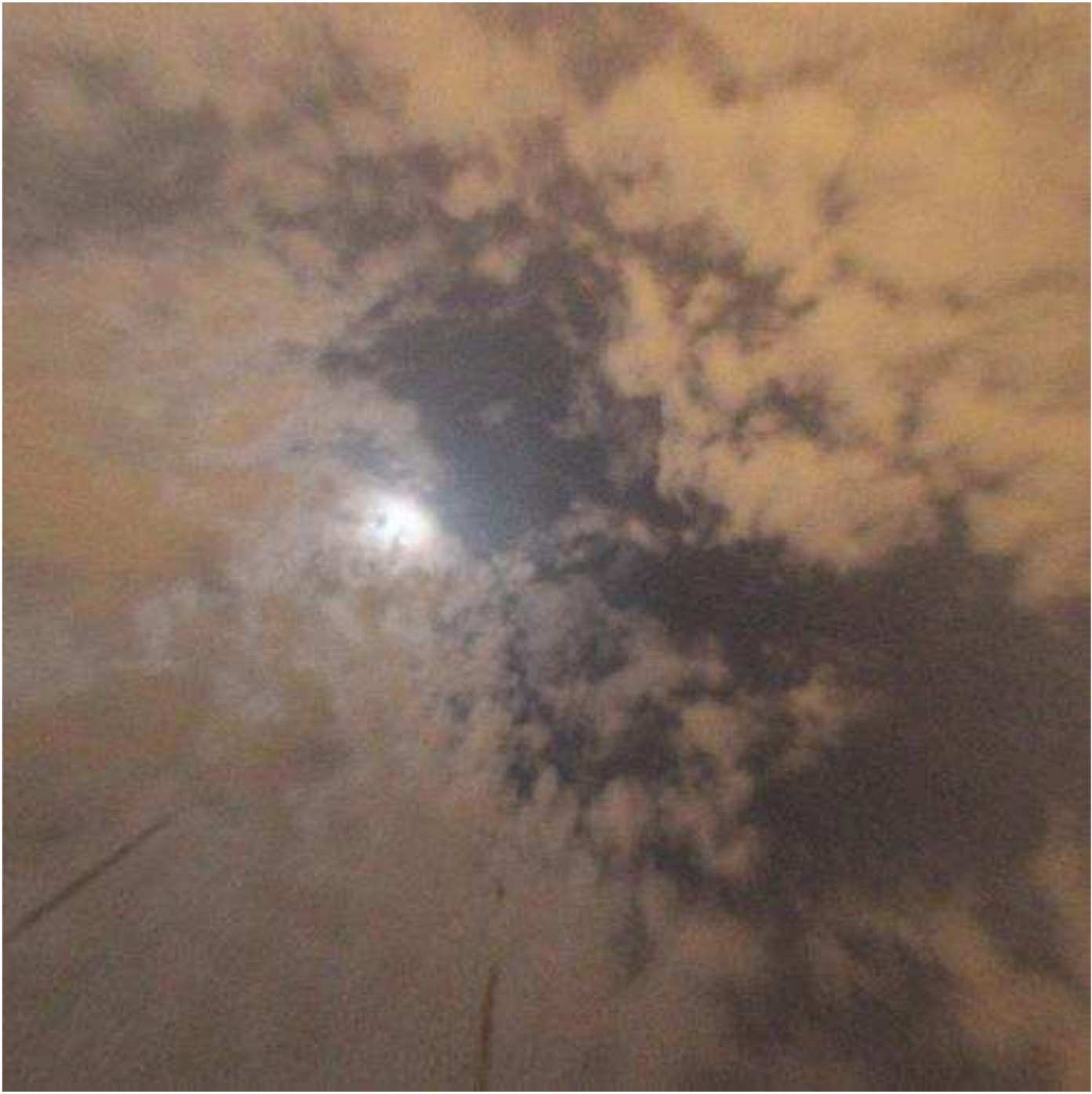}\\
\fbox{\includegraphics[width=0.23\columnwidth]{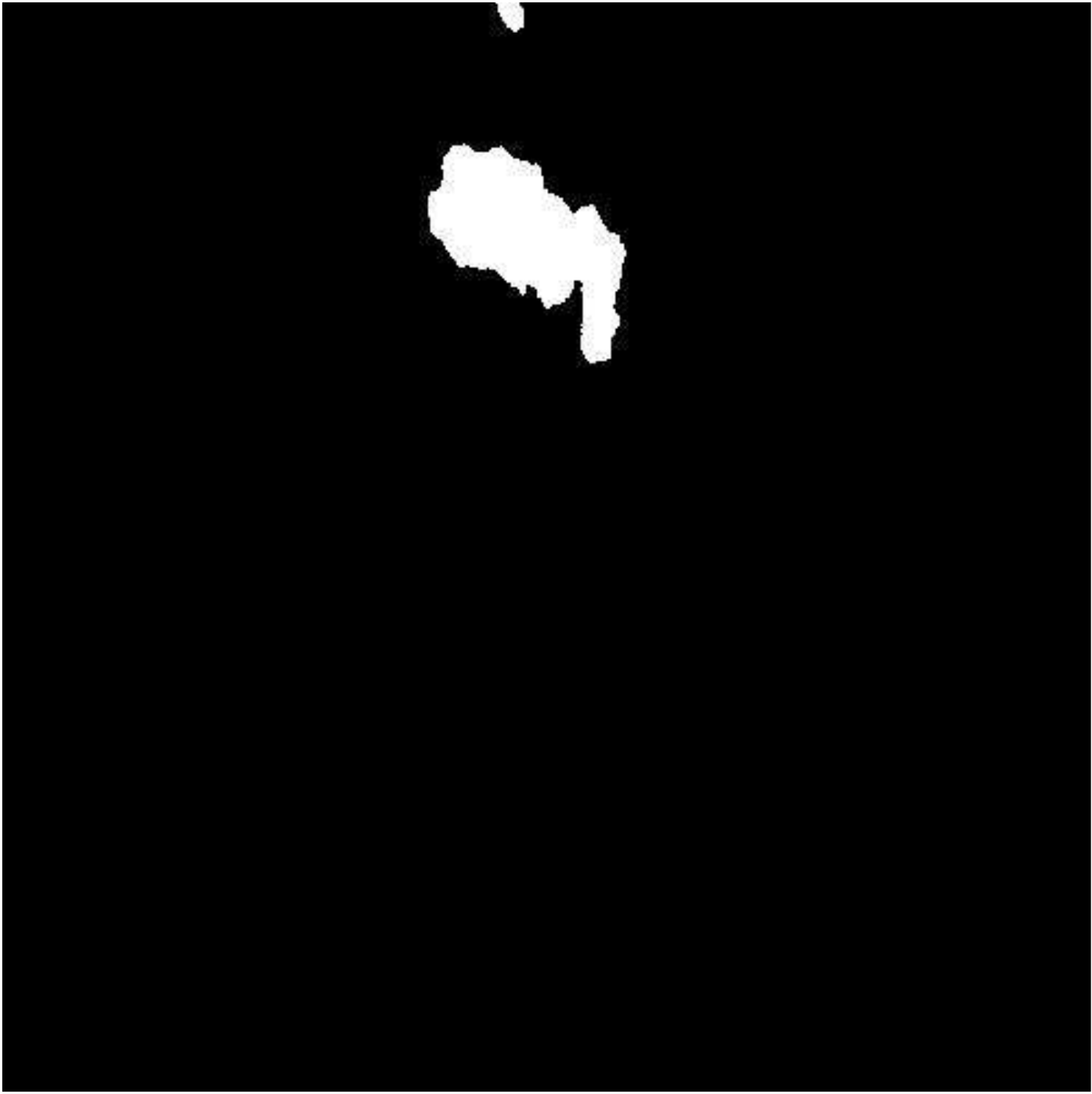}} 
\fbox{\includegraphics[width=0.23\columnwidth]{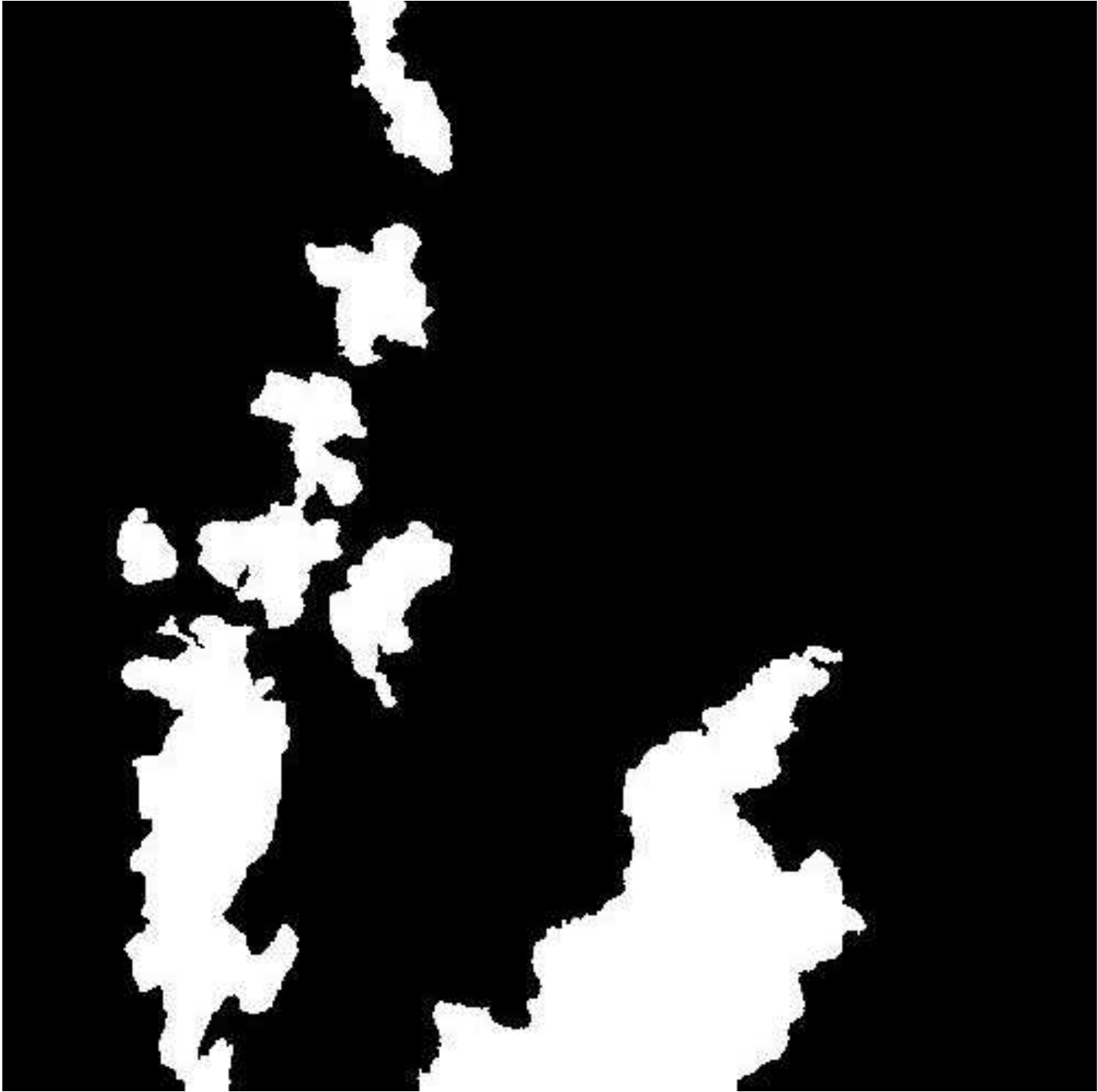}} 
\fbox{\includegraphics[width=0.23\columnwidth]{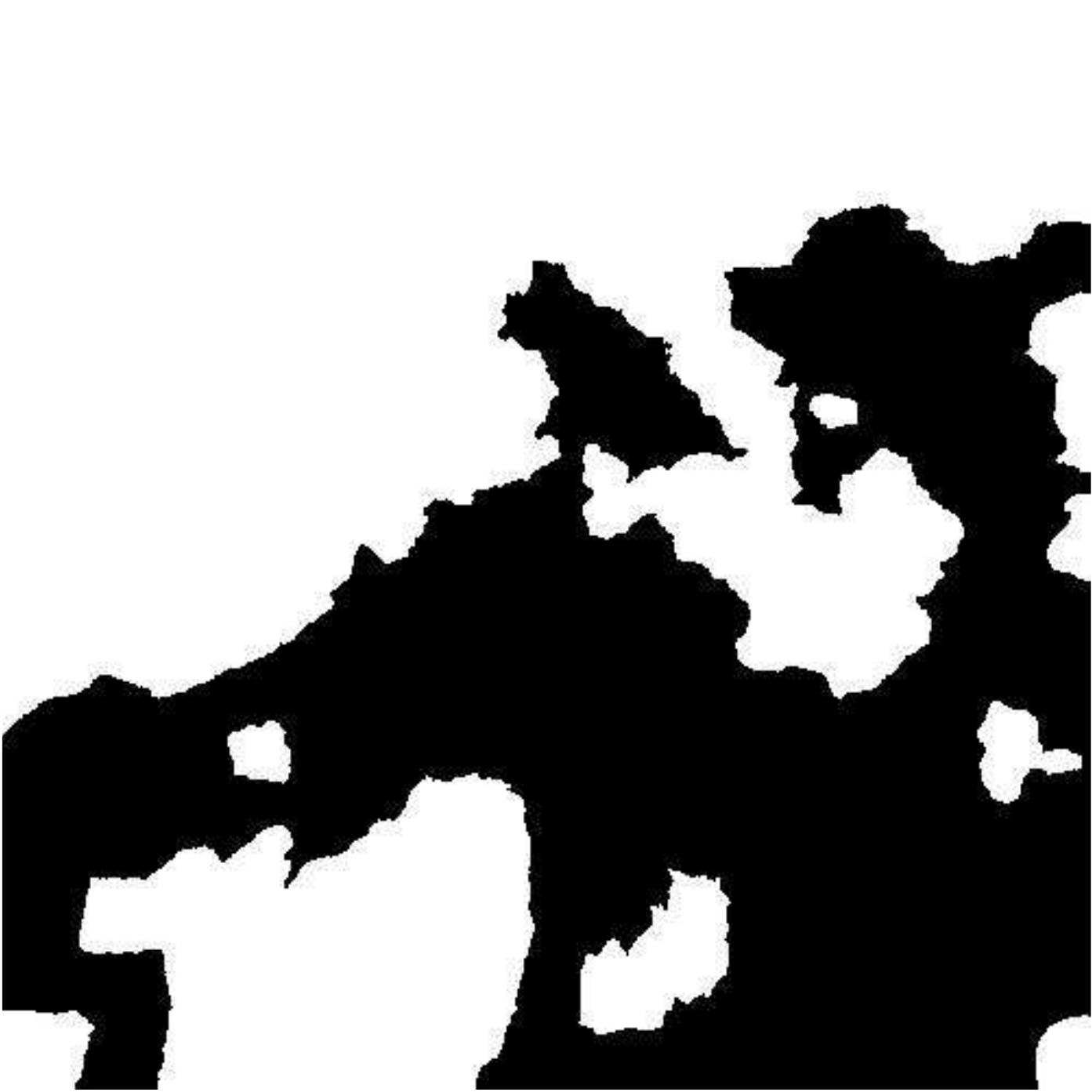}} 
\fbox{\includegraphics[width=0.23\columnwidth]{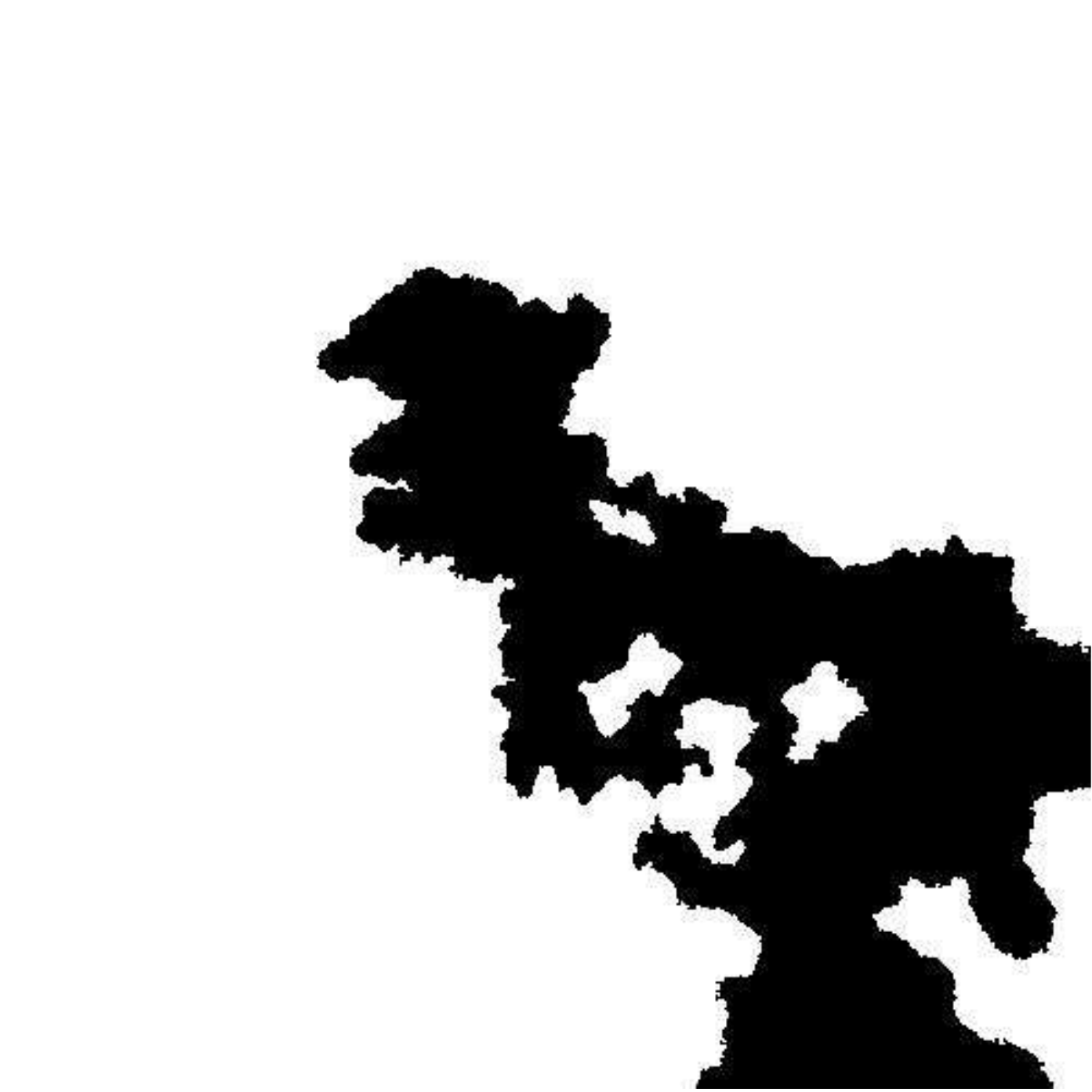}}\\
\makebox[0.11\textwidth][c]{2\%}
\makebox[0.11\textwidth][c]{18\%}
\makebox[0.11\textwidth][c]{60\%}
\makebox[0.11\textwidth][c]{79\%}
\caption{Sample images from the SWINSEG database, along with corresponding sky/cloud segmentation ground truth with varying cloud coverage percentages.}
\label{fig:SWIN-images}
\end{figure}

\begin{figure}[htb]
\centering
\includegraphics[width=0.37\textwidth]{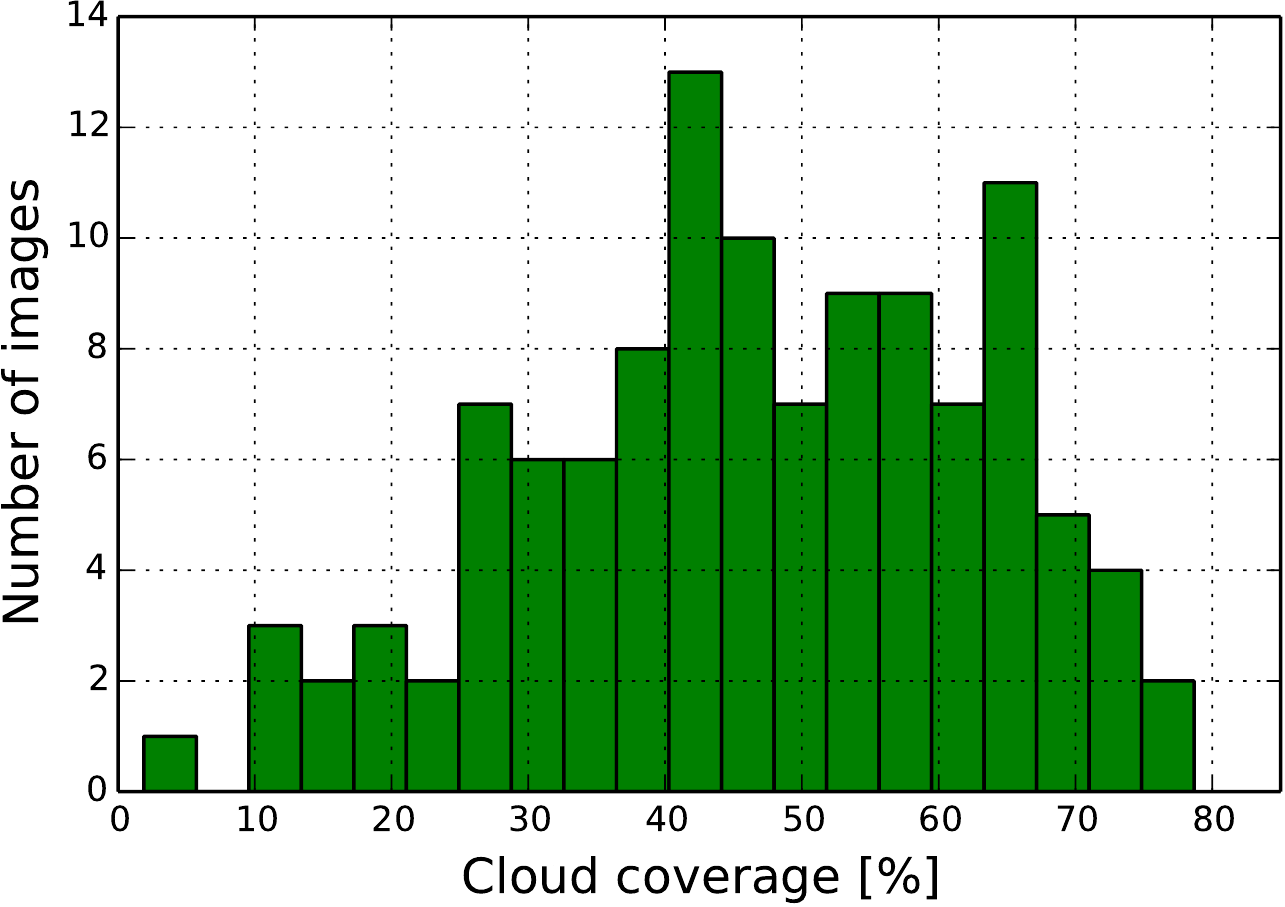}
\caption{Distribution of images in the SWINSEG dataset according to cloud coverage.}
\label{fig:cc-stats}
\end{figure}

SWINSEG consists of $115$ diverse images with a few sample images shown in Fig.~\ref{fig:SWIN-images}. The images were captured using a ground-based whole sky imager called Wide Angle High Resolution Sky Imaging System (WAHRSIS)~\cite{IGARSS2015a}, designed and deployed at Nanyang Technological University in Singapore. The images of SWINSEG were captured during the period January to December 2016. The images were hand-picked such that most of the variation in the image types are captured in the dataset. Several factors such as time of the image capture, cloud coverage, and seasonal variations were considered in the selection process. The distribution of images in SWINSEG  according to cloud coverage percentages is shown in Fig.~\ref{fig:cc-stats}.

\begin{figure}[htb]
\centering
\includegraphics[height=0.2\textwidth]{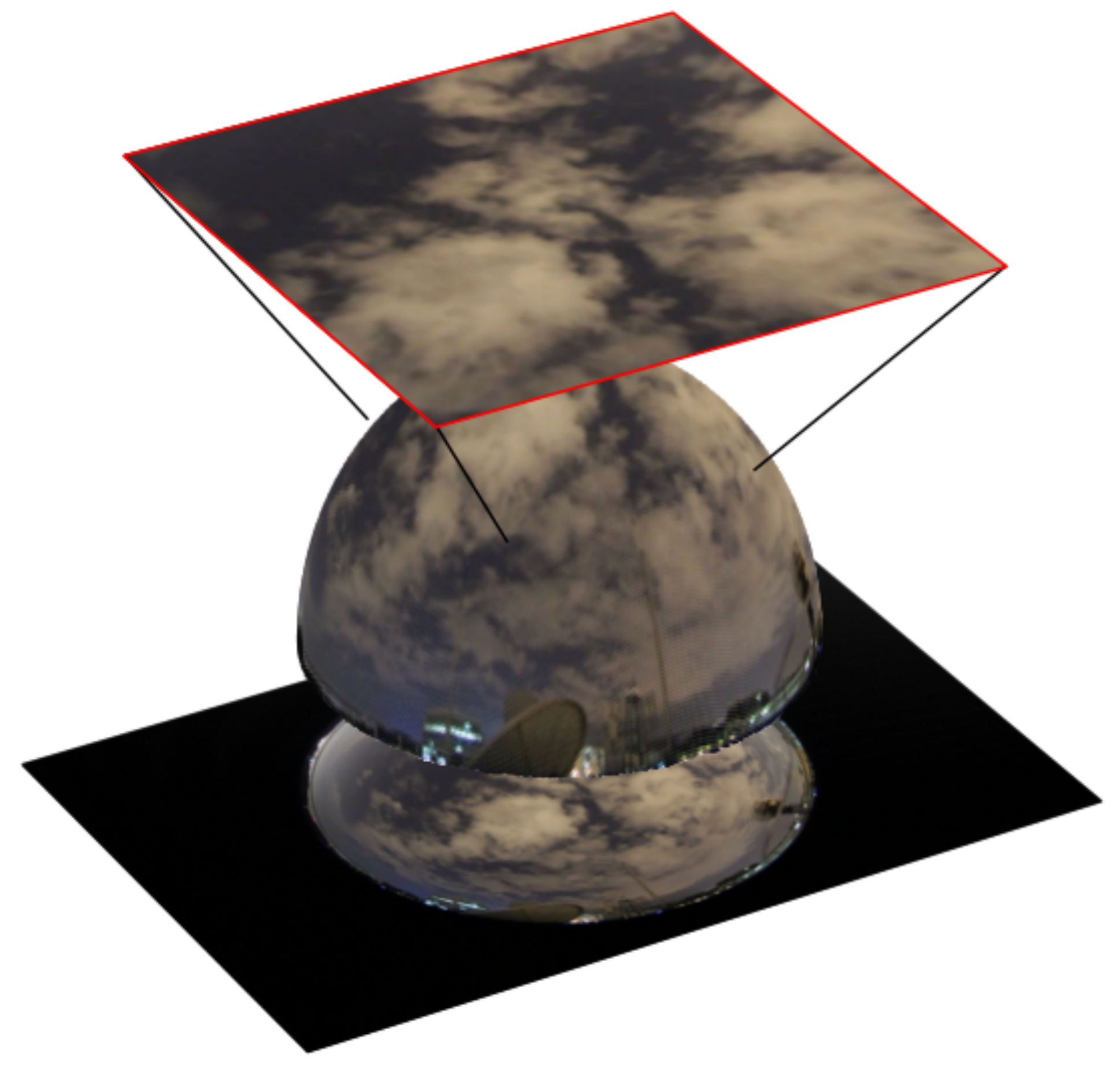}\hspace{0.5mm}    
\includegraphics[height=0.2\textwidth]{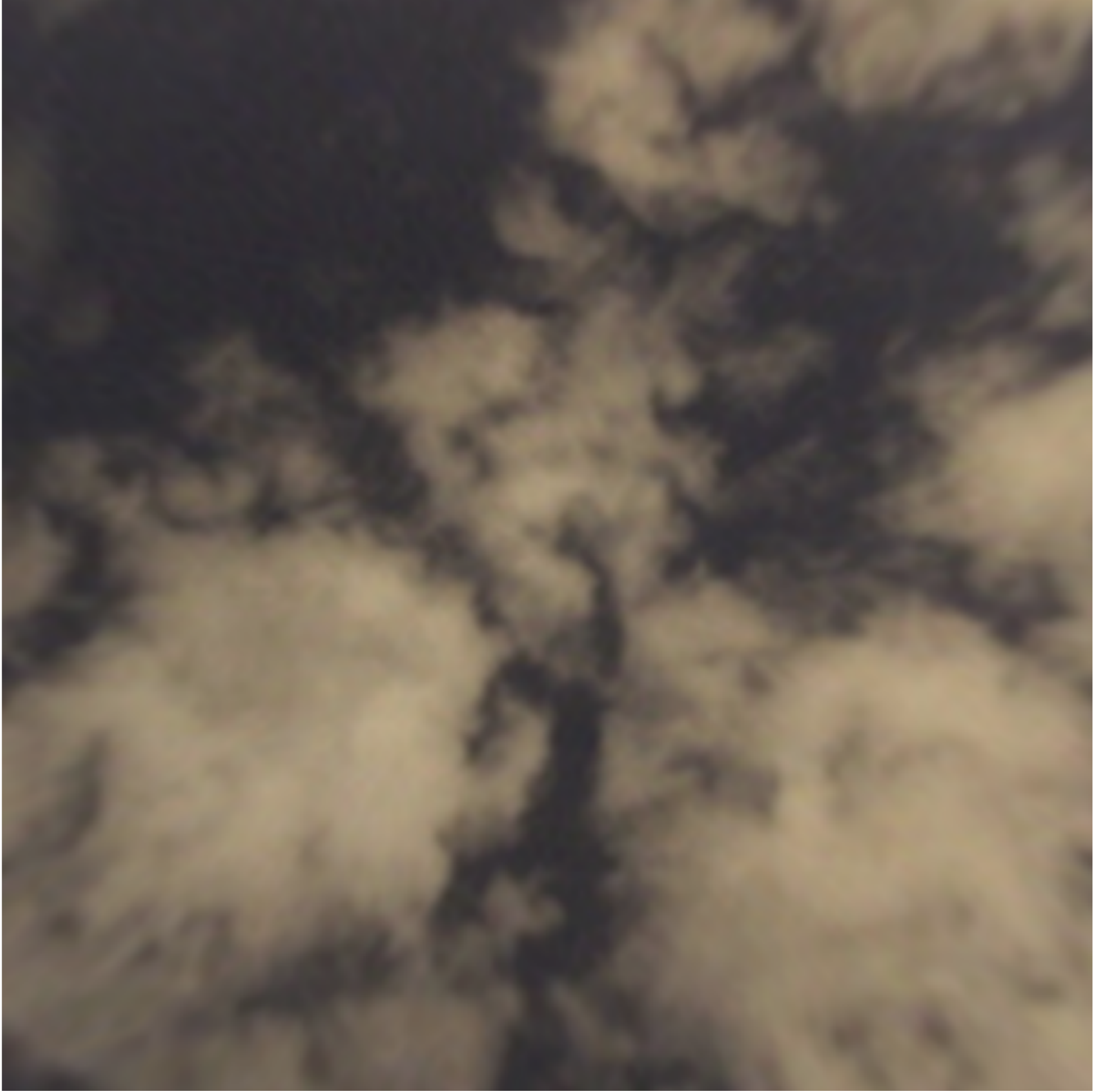}\\
\makebox[0.2\textwidth][c]{(a)}
\makebox[0.2\textwidth][c]{(b)}
\caption{Generation of undistorted images. (a) Illustration of ray-tracing approach with input image at the bottom, projection on the unit sphere, and generation of the undistorted image at the top. (b) Resulting image.}
\label{fig:undistortion-results}
\end{figure}

The extreme wide-angle fish-eye lens of WAHRSIS allows us to capture almost the entire hemisphere.  The distortions introduced by this lens need to be undone before further processing. We generate undistorted pictures of a smaller subsection of the captured images using a ray tracing approach~\cite{JSTARS2016} by considering an imaginary camera (with a standard lens) at the center of the hemisphere, which points towards a user-defined direction. In order to simulate this camera, we consider a target image of dimension $500 \times 500$, with altitude of the virtual image plane at $150$m, where each pixel represents an incident light ray. All those rays will intersect the hemisphere at a certain point and then converge towards the center. The value of a pixel is then equal to the color of the hemisphere at its intersection point. Figure~\ref{fig:undistortion-results} shows the output image as well as the lines corresponding to its borders and diagonals on the original image.

The segmentation ground truth for SWINSEG was generated in consultation with experts from the Singapore Meteorological Service.

\section{Experimental Evaluation}
\label{sec:illust}
In order to present an objective evaluation of the proposed algorithm, we report segmentation accuracy in terms of precision, recall, F-score, and error rate. Suppose \emph{TP}, \emph{TN}, \emph{FP} and \emph{FN} denote the true positive, true negative, false positive and false negative samples of a segmented image, then $\mbox{Precision}=TP/(TP+FP)$ and $\mbox{Recall}=TP/(TP+FN)$. The F-score is the harmonic mean of Precision and Recall. The error rate is the ratio of pixels classified incorrectly.

We also use Receiver Operating Characteristics (ROC) curves to select the optimal color channel for nighttime segmentation. The ROC curve is a graphical plot between true positive rate and false positive rate, as the threshold is gradually changed. We compute the area between the ROC curve and the random classifier slope. These scores are averaged over all images in the dataset. 

\vspace{-0.3cm} 
\subsection{Color Channel Selection}
\label{sec:which-channel}
\begin{figure}[htb]
\centering
\includegraphics[width=0.9\columnwidth]{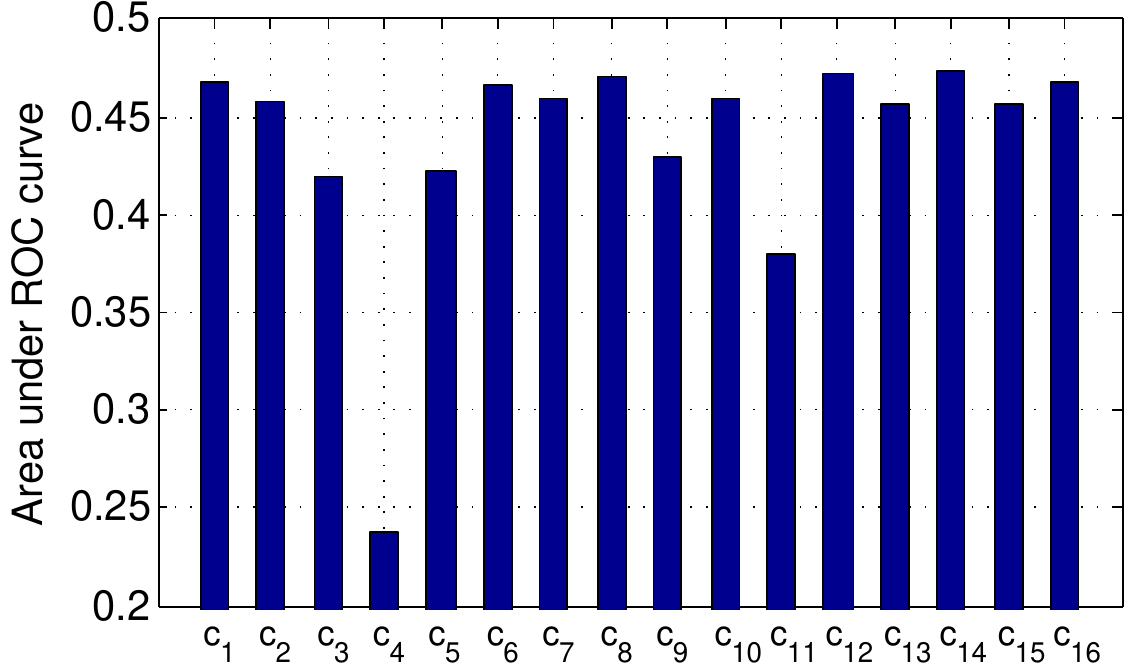}
\caption{Area between ROC curve and random classifier slope for different color channels.}
\label{fig:colorchannels}
\end{figure}

For an efficient nighttime image segmentation, it is important to choose the most discriminatory color channel $\mathbf{c_{*}}$ in our approach. We select the best color channel using the area between the ROC curve and the random classifier slope, to identify the discriminatory capability of the respective color channels. This is shown in Fig.~\ref{fig:colorchannels}. 

The best color channel for nighttime sky/cloud image segmentation is $c_{14}$ ($R-B$), closely followed by $c_{12}$ ($b^{*}$) and $c_8$ ($I$). The ranking of color channels is somewhat different from the one for daytime sky/cloud image segmentation \cite{JSTARS2016}; in particular, those channels involving red are better for night sky/cloud segmentation. 

As discussed in Section~\ref{sec:main}, the over-segmentation is dependent on the number of superpixels $P$ in an image. $P$ should be large enough to capture all the essential details of the sky/cloud image. Ideally each superpixel should contain either only sky or only cloud. We experimentally set $P=100$. This makes the average size of a superpixel for our test images in the SWIMSEG dataset about $50^2$ pixels, which appears to be a good size for this purpose.

\subsection{Benchmarking}

In the literature, thresholding techniques in various color spaces are widely used in cloud segmentation. As discussed in Section~\ref{sec:intro}, we benchmark our proposed framework with several state-of-the-art methods, including Yang et al.\ \cite{Yang2009cloud}, Yang et al.\  \cite{Yang2010}, Liu et al.\ \cite{LiuSP2015} and Gacal et al.\ \cite{Gacal2016}. 

Table~\ref{tab:my-scores}  provides an objective evaluation of these algorithms. We observe that methods such as Gacal et al.\ and Liu et al.\ achieve very high recall but low precision (i.e.\ they are too optimistic about what is a cloud). The method from Yang et al.\ \cite{Yang2009cloud} on the other hand has a high precision and low recall (i.e.\ too optimistic about sky regions). A good segmentation method must have both high precision and high recall, achieving a balance between sky and cloud detection; this is indicated by the F-score and error rate. Our proposed method outperforms other approaches, based on error rates, and has the highest F-score value.

\begin{table}[htb]
\small
\centering
\begin{tabular}{lcccc}
\hline
\textbf{Methods} & \textbf{Precision} & \textbf{Recall} & \textbf{F-score} & \textbf{Error} \\ 
\hline
Yang et al.\ \cite{Yang2009cloud} & \textbf{0.98} & 0.69 & 0.79 & 0.14\\ 
Yang et al.\ \cite{Yang2010} & 0.90 & 0.19 & 0.26 & 0.40\\ 
Liu et al.\ \cite{LiuSP2015} & 0.68 & 0.93 & 0.77 & 0.23\\ 
Gacal et al.\ \cite{Gacal2016} & 0.47 & \textbf{0.99} & 0.62 & 0.53 \\ 
Proposed approach & 0.95 & 0.76 & \textbf{0.83} & \textbf{0.13} \\ 
\hline
\end{tabular}
\caption{Performance evaluation of different benchmarking algorithms. The average scores across all the images of the database are reported. The best performance according to each criterion is indicated in bold.}
\label{tab:my-scores}
\end{table}

\section{Conclusion}
We have presented a robust approach for the segmentation of nighttime sky/cloud images obtained from ground-based sky cameras. Our proposed method is based on superpixel segmentation of the image and is entirely threshold-free. It outperforms other state-of-the-art algorithms. We also release the first nighttime sky/cloud image segmentation database to the research community. Our future work includes increasing the SWINSEG dataset size.

\balance


\begin{thebibliography}{10}
	
	\bibitem{GRSM2016}
	S.~Dev, B.~Wen, Y.~H. Lee, and S.~Winkler,
	\newblock ``Ground-based image analysis: A tutorial on machine-learning
	techniques and applications,''
	\newblock {\em IEEE Geoscience and Remote Sensing Magazine}, vol. 4, no. 2, pp.
	79--93, June 2016.
	
	\bibitem{GRSL2017}
	S.~Dev, F.~M. Savoy, Y.~H. Lee, and S.~Winkler,
	\newblock ``Rough-set-based color channel selection,''
	\newblock {\em IEEE Geoscience and Remote Sensing Letters}, vol. 10, pp.
	52--56, Jan. 2017.
	
	\bibitem{Li2011}
	Q.~Li, W.~Lu, and J.~Yang,
	\newblock ``A hybrid thresholding algorithm for cloud detection on ground-based
	color images,''
	\newblock {\em Journal of Atmospheric and Oceanic Technology}, vol. 28, no. 10,
	pp. 1286--1296, Oct. 2011.
	
	\bibitem{ICIP1_2014}
	S.~Dev, Y.~H. Lee, and S.~Winkler,
	\newblock ``Systematic study of color spaces and components for the
	segmentation of sky/cloud images,''
	\newblock in {\em Proc. International Conference on Image Processing (ICIP)},
	2014, pp. 5102--5106.
	
	\bibitem{ICIP2015a}
	S.~Dev, Y.~H. Lee, and S.~Winkler,
	\newblock ``Multi-level semantic labeling of sky/cloud images,''
	\newblock in {\em Proc. International Conference on Image Processing (ICIP)},
	2015, pp. 636--640.
	
	\bibitem{Yang2009cloud}
	J.~Yang, W.~Lu, Y.~Ma, W.~Yao, and Q.~Li,
	\newblock ``An automatic ground based cloud detection method based on adaptive
	threshold,''
	\newblock {\em Journal of Applied Meteorological Science}, vol. 20, no. 6, pp.
	713--721, 2009.
	
	\bibitem{Yang2010}
	J.~Yang, W.~Lu, Y.~Ma, W.~Yao, and Q.~Li,
	\newblock ``An automatic ground-based cloud detection method based on the local
	threshold interpolation,''
	\newblock {\em Acta Meteorologica Sinica}, vol. 68, no. 6, pp. 1007--1017, Mar.
	2010.
	
	\bibitem{LiuSP2015}
	S.~Liu, L.~Zhang, Z.~Zhang, C.~Wang, and B.~Xiao,
	\newblock ``Automatic cloud detection for all-sky images using superpixel
	segmentation,''
	\newblock {\em IEEE Geoscience and Remote Sensing Letters}, vol. 12, no. 2, pp.
	354--358, Feb. 2015.
	
	\bibitem{Gacal2016}
	G.~F.~B. Gacal, C.~Antioquia, and N.~Lagrosas,
	\newblock ``Ground-based detection of nighttime clouds above {Manila
		Observatory (14.64N, 121.07E)} using a digital camera,''
	\newblock {\em Applied Optics}, vol. 55, no. 22, pp. 6040--6045, Aug. 2016.
	
	\bibitem{Kyba2012}
	C.~C.~M. Kyba, T.~Ruhtz, J.~Fischer, and F.~H\"{o}lker,
	\newblock ``Red is the new black: How the colour of urban skyglow varies with
	cloud cover,''
	\newblock {\em Monthly Notices of the Royal Astronomical Society}, vol. 425,
	no. 1, pp. 701--708, Aug. 2012.
	
	\bibitem{SLIC}
	R.~Achanta, A.~Shaji, K.~Smith, A.~Lucchi, P.~Fua, and S.~S{\"{u}}sstrunk,
	\newblock ``{SLIC} superpixels compared to state-of-the-art superpixel
	methods,''
	\newblock {\em IEEE Transactions on Pattern Analysis and Machine Intelligence},
	vol. 34, no. 11, pp. 2274--2282, Nov. 2012.
	
	\bibitem{Puzicha2000}
	J.~Puzicha, M.~Held, J.~Ketterer, J.~M. Buhmann, and D.~W. Fellner,
	\newblock ``On spatial quantization of color images,''
	\newblock {\em IEEE Transactions on Image Processing}, vol. 9, no. 4, pp.
	666--682, Apr. 2000.
	
	\bibitem{JSTARS2016}
	S.~Dev, Y.~H. Lee, and S.~Winkler,
	\newblock ``Color-based segmentation of sky/cloud images from ground-based
	cameras,''
	\newblock {\em IEEE Journal of Selected Topics in Applied Earth Observations
		and Remote Sensing}, vol. 10, pp. 231--242, Jan. 2017.
	
	\bibitem{IGARSS2015a}
	S.~Dev, F.~M. Savoy, Y.~H. Lee, and S.~Winkler,
	\newblock ``Design of low-cost, compact and weather-proof whole sky imagers for
	{High-Dynamic-Range} captures,''
	\newblock in {\em Proc. International Geoscience and Remote Sensing Symposium
		(IGARSS)}, 2015, pp. 5359--5362.
	
\end{thebibliography}
\end{document}